\documentclass[10pt,twocolumn,letterpaper]{article}

\usepackage{cvpr}
\usepackage{times}
\usepackage{epsfig}
\usepackage{graphicx}
\usepackage{amsmath}

\DeclareMathOperator*{\argmax}{argmax}
\usepackage{mathtools}

\DeclarePairedDelimiter{\floor}{\lfloor}{\rfloor}
\usepackage{upgreek}
\usepackage{graphicx}
\graphicspath{ {./images/} }
\usepackage{booktabs}

\usepackage{amssymb}
\usepackage[numbers]{natbib}
\usepackage[breaklinks=true,bookmarks=false]{hyperref}

\cvprfinalcopy 

\def\cvprPaperID{****} 

\begin{document}

\title{Spatio-temporal Action Recognition: A Survey}

\author{Amlaan Bhoi\\
Department of Computer Science\\
University of Illinois at Chicago\\
{\tt\small abhoi3@uic.edu}
}

\maketitle

\begin{abstract}
   The task of action recognition or action detection involves analyzing videos and determining what action or motion is being performed. The primary subject of these videos are predominantly humans performing some action. However, this requirement can be relaxed to generalize over other subjects such as animals or robots. The applications can range from anywhere between human-computer interaction to automated video editing proposals. When we consider spatio-temporal action recognition, we deal with \textbf{action localization}. This task not only involves determining what action is being performed, but also when and where it is being performed in said video. This paper aims to survey the plethora of approaches and algorithms attempted to solve this task, give a comprehensive comparison between them, explore various datasets available for the problem, and determine the most promising approaches.
\end{abstract}

\section{Introduction}

\textbf{Spatio-temporal action recognition}, or action localization \cite{tian2013spatiotemporal}, is the task of classifying what action is being performed in a sequence of frames (or video) as well as localizing each detection both in space and time. The localization can be visualized using bounding boxes or masks. There has been an increased interest in this task in recent years due to the increased availability of computing resources as well as new advances in convolutional neural network architectures.

There are several approaches to tackle this task. Most of the approaches revolve around the following approaches: discriminative parts \cite{ma2013action, ke2007event}, figure-centric models \cite{klaser2010human, lan2011discriminative, prest2013explicit}, deformable parts \cite{tian2013spatiotemporal}, action proposals \cite{gkioxari2015finding, yu2015fast, jain2014action, weinzaepfel2015learning}, graph-based \cite{yan2018spatial, soomro2015action}, 3D convolutional neural networks \cite{diba2017deep}, and more. We examine each approach, list out advantages and disadvantages, and explore common subset of techniques used between many approaches. We then explore the various datasets available for this task and how they are a sufficient metric to evaluate this problem. Finally, we comment on which methods are promising going forward. The term spatio-temporal action recognition and action localization will be used interchangeably in this text as they refer to the same task. We should not confuse action localization with the similarly framed problem of \textbf{temporal action detection} which deals with determining only \textit{when} an action occurs in a large video.

\section{Problem Definition}

We can broadly define the problem as: given a video $X = \{x_1, x_2, ..., x_n\}$ where $x_i$ is the $i^{th}$ frame in the video, determine the action label $a_i \in A$, where $A$ is the set of action labels in the dataset, for that $i^{th}$ frame as well as a set of $\{x1, x2, y1, y2\}$ coordinates of the bounding box of the classified action $a_i$.

An alternative formulation given by \citet{weinzaepfel2015learning} defines action localization as: given a video of T frames $\{I_t\}_{t=1..T}$ and a class $c \in C$ where C is the set of classes, the task involves detecting if action $c$ occurs in the video and if yes, when and where. The output of a successful algorithm should output $\{R_t\}_{t=t_b..t_e}$ with $t_b$ the beginning and $t_e$ the end of the predicted temporal extent of action $c$ and $R_t$ the detected region in frame $I_t$.

Every paper contains a slightly different formulation of the problem depending on the approach they have taken. We shall swiftly explore those definitions to see how this one task can be approached in different viewpoints (graphs, optical flow, etc).

\section{Challenges}

Spatio-temporal action recognition faces the usual challenges in faced in action recognition such as tracking the action throughout the video, localizing the time frame when the action occurs, and more. However, there are an additional set of challenges such as but not limited to:

\begin{itemize}
    \item Background clutter or object occlusion in video
    \item Spatial complexity in scene with respect to number of candidate objects
    \item Linking actions between frames in presence of irregular camera motion
    \item Predicting optical flow of action
\end{itemize}

However, there is a more fundamental problem to consider with the traditional approach to action localization. We cannot treat the problem in a linear way of just classifying an action. Even object detection algorithms require region proposals to classify \cite{ren2015faster}. This can be made worse by the introduction of the temporal dimension. This would cause an exponential increase in the number of proposals which would render any such approach impractical for use.

\section{Action Proposal Models}

\subsection{Action localization with tubelets from motion}

\citet{jain2014action} propose a method for spatio-temporal action recognition by proposing a selective search sampling strategy for videos. Their approach uses \textbf{super-voxels} instead of super-pixels. In this way, they directly obtain the $2D+t$ sequences of bounding boxes which is called as \textit{tubelets}. This removes the issue of linking bounding boxes between frames in a video. In addition to this, their method explicitly incorporates motion information by introducing \textit{independent motion evidence} as a feature to characterize how the action's motions deviates from background motion.

The pipeline of this method starts with super-voxel segmentation which is done through a graph-based method \cite{xu2012evaluation}, iterative generation of additional tubelets, descriptor generation (BOW representation), and finally classification using BOW histograms of tubelets.

\textbf{Super-voxel generation. } Initial super-voxels are agglomeratively merged together based on similarity measures. We can imagine the merging as a tree with the individual super-voxels being the leaf of the tree and being merged all the way up to the root. This procedure produces $n-1$ additional super-voxels.

\textbf{Tubelets}
Wherever super-voxels appear, they are tightly bounded by a rectangular bounding box. A sequence of these bounding boxes produces what is known as a \textit{tubelet}. The algorithm, thus, produces $2n-1$ tubelets.

\subsubsection{Merging}
\begin{figure*}[t]
    \centering
    \includegraphics[width=0.95\linewidth]{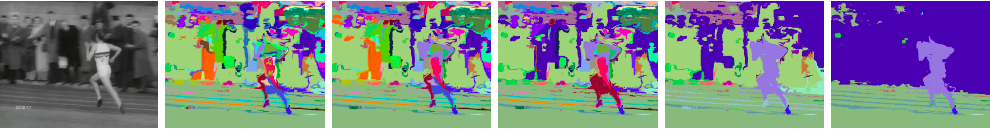}
    \caption{An example of \textbf{Running} action class where the first two images depict a video frame and initial super-voxel segmentation. The other four images represent segmentation after various merge operations.}
    \label{fig:4-1}
\end{figure*}

Merging of super-voxels is based on five criterias that are sectioned into two parts: \textbf{color, texture, motion:}
\begin{equation} \label{eq: 1}
    h_t = \frac{\Gamma (r_i) \times h_i + \Gamma (r_j) \times h_j}{\Gamma (r_i) + \Gamma (r_j)}
\end{equation}
where $h_i$ is the $\ell_1$-normalized histogram for super-voxel $r_i$ (similarly for $r_j$) and $\Gamma (r_i)$ is the number of super-voxels in $r_i$.
The second part is \textbf{size, fill:}
\begin{equation} \label{eq: 2}
    s_{\Gamma}(r_i, r_j) = 1 - \frac{\Gamma (r_i) + \Gamma (r_j)}{\Gamma (video)}
\end{equation}
where $\Gamma (video)$ is the size of the video (in pixels). The merging strategies can vary with any number of combinations of these criterias. An example of the merge operations is illustrated in Figure \ref{fig:4-1}.

\subsubsection{Motion features}
The authors defined \textit{independent motion evidence} (IME) as:
\begin{equation} \label{eq: 3}
    \xi (p,t) = 1 - \varpi (p)
\end{equation}
where $\varpi(p)$ is the ratio $\frac{\phi(r_{\hat{\theta}}(p, t))}{r_{\hat{\theta}}(p, t)}$ normalized between [0, 1]. More details are available in their paper.

\subsubsection{Results} The authors evaluated their model with ROC curve comparisons with other methods. The graphs can be found in their paper. They also recorded mean precision for the MSR-II dataset \cite{zhang2016large} with results shown in Table \ref{table:4-1}.

\begin{table}[ht]
\centering
\label{tubelets}
\begin{tabular}{|c|c|c|c|}
\hline
Method     & Boxing        & Handclapping  & Handwaving    \\ \hline
Cao \textit{et al.} & 17.5          & 13.2          & 26.7          \\ \hline
SDPM       & 38.9          & 23.9          & 44.7          \\ \hline
Tubelets   & \textbf{46.0} & \textbf{31.4} & \textbf{85.8} \\ \hline
\end{tabular}
\caption{Results for \citet{jain2014action}. Average precisions for MSR-II.}
\label{table:4-1}
\end{table}

\subsection{Tube Convolutional Neural Network (T-CNN) for Action Detection in Videos}

\citet{hou2017tube} introduce a new architecture called \textbf{tube convolutional neural networks} or T-CNN which is a generalization of R-CNN \cite{girshick2014rich} from 2D to 3D. Their approach first divides the videos into clips with 8 frames in each clip. This allows them to use a fixed-sized ConvNet architecture to process clips while mitigating the cost of GPU memory. As an input video is processed clip by clip, action tube proposals with various spatial and temporal sizes are generated for various clips. They need to be linked into a tube proposal sequence. 

The authors introduce a new layer called \textbf{Tube-of-Interest} (ToI) pooling layer. This is a 3D generalization of Region of Interest (RoI) pooling layer of R-CNN. ToI layers are used to produce fixed length feature vectors which solves the problem of variable length feature vectors. More details about ToI layers can be found in there paper.

\begin{figure*}[t]
    \centering
    \includegraphics[width=1.0\linewidth]{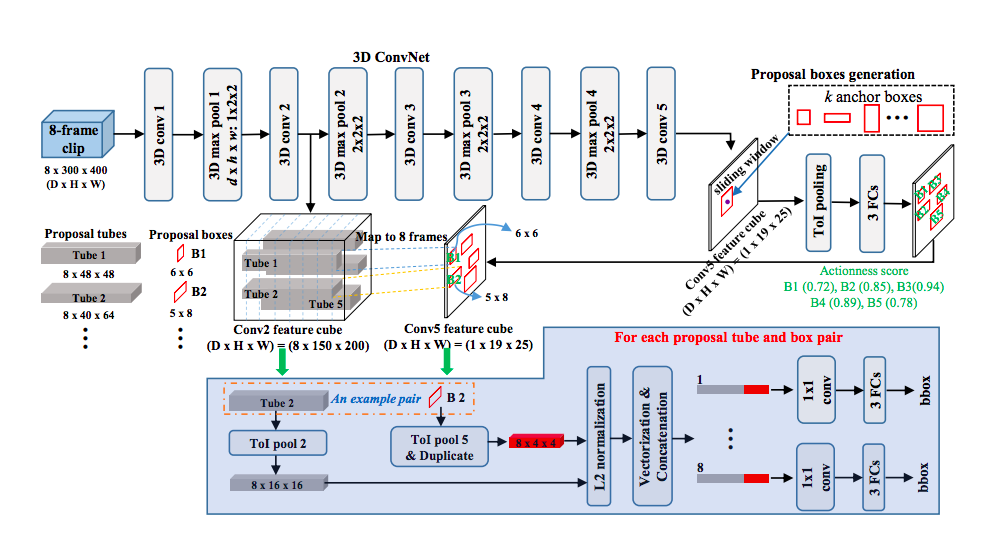}
    \caption{Tube proposal network with all modules.}
    \label{fig:4-2}
\end{figure*}

\subsubsection{Tube Proposal Network}
The TPN consists of 8 3DConv layers, 4 max-pooling layers, 2 ToI layers, 1 point-wise convolutional layer, and 2 fully-connected layers. The authors use a pre-trained C3D model \cite{tran2015learning} and fine-tune it for each dataset they used in experiments. When generalizing R-CNN, instead of using the 9 hand-picked anchor boxes, they decided to use K-Means clustering on training set to select 12 anchor boxes which is better adapted to different datasets. Each bounding box is assigned an "actionness" probability determining if a box corresponds to an action or not (binary label). A positive bounding box proposal is determined by an Intersection-over-Union (IoU) overlap of more than 0.7.

\subsubsection{Linking Tube Proposals}
The primary problem of linking tube proposals is not every consecutive tube proposal may capture the entire action (think of occlusion or noise in middle clips). To solve this, the authors used two metrics when linking tube proposals: actionness and overlap scores. Each video proposal (link of tube proposals) is assigned a score defined as:
\begin{equation} \label{eq: 4}
    S = \frac{1}{m}\sum_{i=1}^{m}Actionness_{i}+\frac{1}{m-1}\sum_{j=1}^{m-1}Overlap_{j, j+1}
\end{equation}
where $Actionness_{i}$ denotes the actionness score of tube proposal from $i$-th clip, $Overlap_{j, j+1}$ measures the overlap between two linked proposals from clips $j$ and ($j$ + 1), and $m$ is the total number of clips.

\subsubsection{Action Detection}
The input to the action detection module is a set of linked tube proposal sequences of varying lengths. This is where the ToI layer comes into action. The output of the ToI layer is atteched to two fully-connected layers and a dropout layer. The dimension of the last fully-connected layer is $N$ + 1 ($N$ action classes and 1 background class).

\subsubsection{Results}
\citet{hou2017tube} evaluated and verified their approach on three trimmed video datasets (UCF-Sports \cite{rodriguez2008action}, J-HMDB \cite{jhuang2013towards}, UCF-101 \cite{soomro2012ucf101}) and one un-trimmed video dataset -- THU-MOS'14 \cite{jiang2014thumos}. The results for the UCF-Sports dataset is outlined in Table \ref{table:4-2} 

There are many more approaches related to action proposal models that we did not get into detail of. These include \textbf{Finding Action Tubes} by \citet{gkioxari2015finding}, \textbf{Fast action proposals for human action detection and search} by \citet{yu2015fast}, \textbf{Learning to track for spatio-temporal action localization} by \citet{weinzaepfel2015learning}, and \textbf{Human Action Localization with Sparse Spatial Supervision} by \citet{weinzaepfel2017human}.

\begin{table*}[t]
\centering
\small
\begin{tabular}{|l|l|l|l|l|l|l|l|l|l|l|l|}
\hline
                           & Diving         & Golf           & Kicking        & Lifting        & Riding          & Run            & SkateB.        & Swing          & SwingB.        & Walk           & mAP           \\ \hline
\citet{weinzaepfel2015learning} & 60.71          & 77.55          & 65.26          & \textbf{100.0} & 99.53           & 52.60          & 47.14          & \textbf{88.88} & 62.86          & 64.44          & 71.9          \\ \hline
\citet{peng2016multi}        & \textbf{96.12} & 80.47          & 73.78          & 99.17          & 97.56           & 82.37          & 57.43          & 83.64          & 98.54          & 75.99          & 84.51         \\ \hline
\citet{hou2017tube}         & 84.38          & \textbf{90.79} & \textbf{86.48} & 99.77          & \textbf{100.00} & \textbf{83.65} & \textbf{68.72} & 65.75          & \textbf{99.62} & \textbf{87.79} & \textbf{86.7} \\ \hline
\end{tabular}
\caption{Results for \citet{hou2017tube}. mAP for each class of UCF-Sports. The IoU threshold $\alpha$ for frame m-AP is fixed to 0.5.}
\label{table:4-2}
\end{table*}

\section{Figure-Centric Models}



\subsection{Discriminative figure-centric models for joint action localization and recognition}

\citet{lan2011discriminative} approach spatio-temporal action recognition by combining bag-of-words style statistical representation and a figure-centric structural representation which mainly works like template matching. They treat the position of the human as a latent variable in a \textbf{discriminative latent variable model} and infer it while simultaneously recognizing an action. In addition, instead of simple bounding boxes, they actually learn discriminative cells within boxes for more robust detection. Due to the latent variable model, exact learning and inference is intractable. Thus, efficient approximate learning and inference algorithms are developed.

\begin{table}[ht]
\centering
\begin{tabular}{@{}ll@{}}
\toprule
Method                    & \multicolumn{1}{c}{Accuracy} \\ \midrule
global bag-of-words       & \multicolumn{1}{c}{63.1}     \\
local bag-of-words        & \multicolumn{1}{c}{65.6}     \\
spatial bag-of-words with $\Delta_{0/1}$ & \multicolumn{1}{c}{63.1}     \\
spatial bag-of-words with $\Delta_{joint}$ & \multicolumn{1}{c}{68.5}     \\
latent model with $\Delta_{0/1}$  & \multicolumn{1}{c}{63.7}     \\
\citet{lan2011discriminative}                       & \multicolumn{1}{c}{73.1}     \\ \bottomrule
\end{tabular}
\caption{Results for \citet{lan2011discriminative}. Mean per-class action recognition accuracies.}
\label{Table:5-1}
\end{table}

\subsubsection{Figure-Centric Video Sequence Model}
The model jointly learns the relationship between action label and location of the person performing the action in each frame. The standard bounding box is divided into $R$ cells where each cell is either turned "on" or "off" depending on if it contains an action.

Each video \textbf{I} has an associated label $y$. Suppose video \textbf{I} contains $\tau$ frames represented as $\textbf{I} = (I_1, I_2,...,I_\tau)$, where $I_i$ denotes the $i$-th frame of said video. Furthermore, the authors define the bounding box for each video as $L = (l_1, l_2,...,l_\tau)$. The $i$-th bounding box $l_i$ is a 4-dimensional tensor representing $(x, y)$ coordinates, height, and width of bounding box. The extracted feature vector $\lambda(l_i;I_i)$ is a concatenation of three vectors: i.e. $\lambda(l_i;I_i) = [\textbf{x}_i;\textbf{g}_i;c_i]$. $\textbf{x}_i$ and $\textbf{g}_i$ represent the appearance feature which is defined as the k-means quantized HOG3D descriptors \cite{klaser2010learning} and spatial locations of interest in bounding box $l_i$. $c_i$ denotes the color histogram.

The authors use a scoring function inspired by the latent SVM model \cite{yu2009learning} which is defined as:

\begin{equation} \label{eq: 5}
    \begin{split}
    \theta^{\top}\Phi(\textbf{z}, L, y, \textbf{I}) = & \sum_{i \in \nu}\alpha^{\top}\phi(l_i, \textbf{z}_i, y, I_i) \\
    & + \sum_{i, i+1 \in \varepsilon}\beta^{\top}\psi (l_i, l_{i+1}, \textbf{z}_i, \textbf{z}_{i+1}, I_i, I_{i+1}) \\
    & + \gamma^{\top}\eta (y, \textbf{I})
    \end{split}
\end{equation}

The definition of the \textbf{unary} potential function, \textbf{pairwise} potential function, and \textbf{global action} potential function can be found in the paper.

\subsubsection{Learning}
Given N training samples $\langle\textbf{I}^n, L^n, y^n\rangle(n = 1, 2,...,N)$, the authors optimize over model parameter $\theta$. They adopt the SVM framework for learning as:

\begin{equation}
    \begin{aligned}
    \min_{\theta, \xi \geq 0} \frac{1}{2}\left \| w \right \|^{2}+C \sum_{n=1}^{N}\xi^n \\
    \textup{s.t.} f_{\theta}(y^{n}, L^n, \textbf{I}^n) - f_{\theta}(y, L, \textbf{I}^n) \geq \\
    \Delta(y, y^n, L, L^n) - \xi^n, \forall n, \forall y, \forall L
    \end{aligned}
\end{equation}

\subsubsection{Inference} 
The inference is simply solving the following optimization problem:

\begin{equation}
    \max_y \max_L f_{\theta}(L, y, \textbf{I}) = \max_{y} \max_{L} \max_{\textbf{z}} \theta^\top \Phi (\textbf{z}, L, y, \textbf{I})
\end{equation}

For a fixed $y \in \mathcal{Y}$, we can maximize $L$ and \textbf{z} as:

\begin{equation}
    \max_L \max_{\textbf{z}} \theta^\top \Phi(\textbf{z}, L, y, \textbf{I})
\end{equation}

\subsubsection{Results}
The authors evaluate their model on the UCF-Sports dataset \cite{rodriguez2008action}. They achieved a \textbf{83.7}\% accuracy beating out previous methods. The complete results for with mean per-class action recognition accuracies can be found in Table \ref{Table:5-1}

The other two closely related Figure-Centric approaches are \textbf{Human focused action localization in video} by \citet{klaser2010human} and \textbf{Explicit modeling of human-object interactions in realistic videos} by \citet{prest2013explicit}.



\section{Deformable Parts Models}

\subsection{Action recognition and localization by hierarchical space-time segments}

\begin{figure*}[ht]
    \centering
    \includegraphics[width=1.0\linewidth]{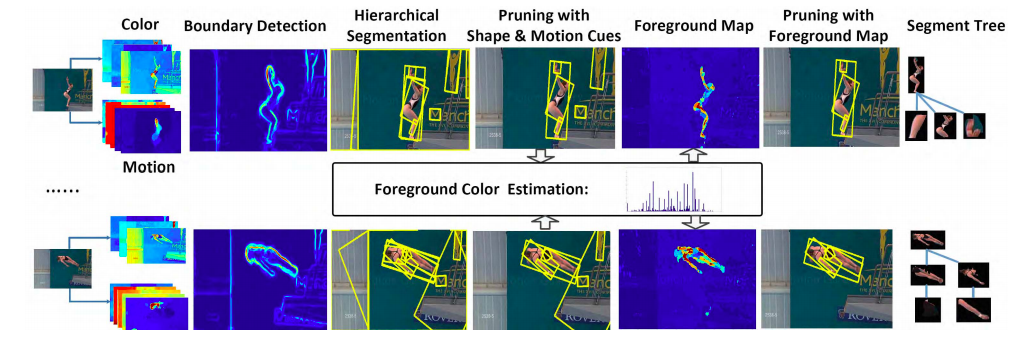}
    \caption{Pipeline for hierarchial video frame segments extraction. \cite{ma2013action}}
    \label{fig:6-1}
\end{figure*}

\citet{ma2013action} introduce a new representation called \textit{hierarchial space-time segments} where the space-time segments of videos are organized into two-level hierarchy. The first level comprises the root space-time segments that may contain the whole human body. The second level comprises space-time segments that contain parts of the root. They present an unsupervised algorithm designed to extract time segments that preserve both static \textit{and} non-static relevant space time segments as well as their hierarchial and temporal relationships. Their algorithm consists of three major steps:
\begin{enumerate}
	\item Apply hierarchical segmentation on each video frame to get a set of segment trees, each of which is considered as a candidate segment tree of the human body.
	\item Prune the candidates by exploring several cues such as shape, motion, articulated objects’ structure and global foreground color.
	\item Track each segment of the remaining segment trees in time both forward and backward.
\end{enumerate}

Finally, using a simple linear SVM on the \textit{bag of hierarchial space-time segments} representation, they achieved better or comporable perforamnce on previous methods.

\begin{figure*}[ht]
    \centering
    \includegraphics[width=1.0\linewidth]{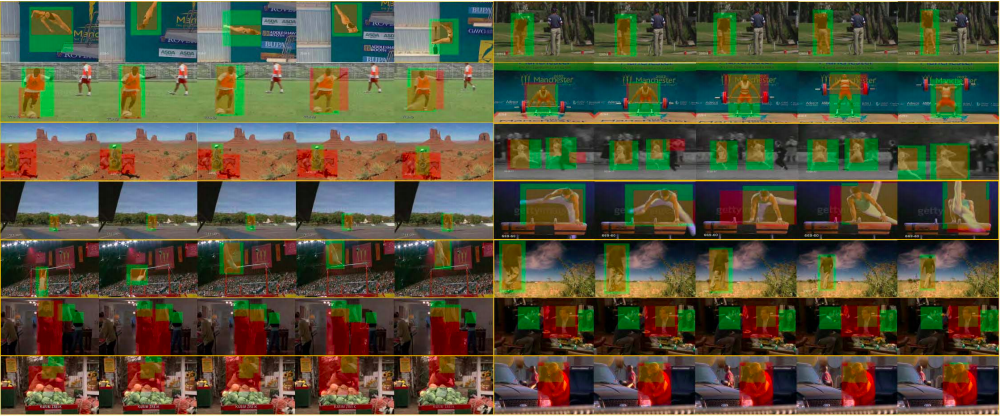}
    \caption{Example of action localization results on UCF-Sports \cite{rodriguez2008action} and High-Five \cite{patron2010high} datasets. \cite{ma2013action}}
    \label{fig:6-1-1}
\end{figure*}

\subsubsection{Video Frame Hierarchial Segmentation}
On each video frame, the authors compute the boundary map as described by \cite{leordeanu2012efficient} using three color channels and five motion channels including optical flow, unit normalized optical flow, and optical flow magnitude. The boundary map is then utilized to compute Ultrametic Contour Map (UCM) as described by \cite{arbelaez2009contours}. By traversing the UCM, certain segments are removed to reduce redundancy. Then, the authors remove the root of the segment tree and get a set of segment trees $\tau^{t}$ where $t$ is the frame index. Each $T_{j}^{t} \in \tau^{t}$ is considered a candidate segment tree of a human body and we denote $T_{j}^{t} = {s_{ij}^{t}}$ where each ${s_{ij}^{t}}$ is a segment and ${s_{0j}^{t}}$ is the root segment.

\subsubsection{Pruning Candidate Segment Trees}
The pruning step should only prune irrelevant static segments. The decision to prune a candidate segment is made with information from all segments of the tree and not just local information. Thus, pruning is done at candidate level. The two methods used to perform tree pruning are with \textbf{shape and color cues} and \textbf{using foreground map}. Detailed representations are available in the original paper.

\subsubsection{Extracting Hierarchial Space-Time Segments}
After pruning, there is a set $\hat\tau^{t}$ containing remaining candidate segment trees. To capture temporal information, for each $T_{j}^{t} \in \hat\tau^{t}$, the authors track every segment $s_{ij}^{t} \in T_{j}^{t}$ to construct a space-time segment. The authors then propose a non-rigid region tracking method. The method revolves around predicting the region of next frame by flow and computing a flow prediction map $M_f$ as well as color prediction map $M_c$. If a point $b' \in B$ (where $B$ is the bounding box of next frame inside region $R'$) has color $c_{b'}$, then $M_{c}(b') = \textbf{h}(c_{b'})$ and $M_f(b')$:

\begin{equation}
    M_{f}(b') = \begin{cases}
    2 & \text{$b' \in R'$} \\
    1 & \text{$b' \in \hat{B} \wedge b' \notin R'$} \\
    0 & \text{otherwise}
  \end{cases}
\end{equation}

The combined map $M$ is then scaled and quantized to contain integer values in range $[0, 20]$. By settings thresholds $\delta_m$ of integer values between 1 to 20, we get 20 binary maps. The size of every connected component is computed and one with most similar size to $R$ is selected as candidate. These space-time segments may contain same objects. To exploit this dense representation, we can group space-time segments together if they overlap over some threshold (authors used 0.7). Finally, for each track (groups of segments), bounding boxes are calculated on all spanned frames.

\subsubsection{Action Recognition and Localization}
For action recognition, a one-vs-all linear SVM is trained on all training videos' BoW representations for multiclass classification resulting in the following rule:

\begin{equation}
    y = \argmax_{y \in \mathcal{Y}}
    \left( \begin{array}{ccc}
            \textbf{w}_{y}^{r} \\
            \textbf{w}_{y}^{p}\end{array} \right)
    (\textbf{x}^{r}\textbf{x}^{p})+b_y
\end{equation}
where $\textbf{x}^{r}$ and $\textbf{x}^{p}$ are BoW representations of root and part space-time segments of test video respectively, $\textbf{w}_{y}^{r}$ and $\textbf{w}_{y}^{p}$ and entries of trained separation hyperplane for roots and parts respectively, $b_y$ is the bias term, and $\mathcal{Y}$ is set of action class labels.

For action localization, we find space-time segments that have positive contribution to classification of the video. Given a test video and set of root space-time segments $S^{p} = {\textbf{s}_{a}^{p}}$ and set of part space-time segments $S^{p} = {\textbf{s}_{b}^{p}}$, denote $C^r = {\textbf{c}_{k}^{r}}$ and $C^p = {\textbf{c}_{k}^{p}}$ as set of code words corresponding to positive entries of $\textbf{w}_{y}^{r}$ and $\textbf{w}_{y}^{p}$ respectively. We compute set U as:

\begin{equation}
    \begin{aligned}
    U = \{ \hat{s}^{r} : \hat{s}^{r} = \argmax_{s_{a}^{r} \in S^{r}} h(\textbf{s}_{a}^{r}, \textbf{c}_{k}^{r}), \forall \textbf{c}_{k}^{r} \in C^{r} \} \\ 
    \cup \{ \hat{s}^{p} : \hat{s}^{p} = \argmax_{s_{b}^{p} \in S^{p}} h(\textbf{s}_{b}^{p}, \textbf{c}_{k}^{p}), \forall \textbf{c}_{k}^{p} \in C^{p} \}
    \end{aligned}
\end{equation}
where function $h$ measures similarity between two space-time segments. Finally, the tracks are output which have at least one space-time segment in set $U$ as action localization results.

\subsubsection{Results}
\citet{ma2013action} experimented on the UCF-Sports \cite{rodriguez2008action} and High Five \cite{patron2010high} datasets. In action localization performance, they had an average of 10\% increase in average IOU compared to previous methods.

\begin{table}[ht]
\centering
\scriptsize
\begin{tabular}{@{}|l|c|c|c|c|c|c|c|c|@{}}
\toprule
\multicolumn{4}{c|}{subset of frames}                    & \multicolumn{4}{c|}{all frames}                     \\ \cmidrule(l){2-9} 
                  & {\cite{tran2011optimal}} & {\cite{tran2012max}}      & {\cite{lan2011discriminative}}      & Ma            & {\cite{tran2011optimal}} & {\cite{tran2012max}}      & {\cite{lan2011discriminative}} & Ma            \\ \midrule
dive              & 16.4     & 36.5          & 43.4          & \textbf{46.7} & 22.6     & 37.0          & -        & \textbf{44.3} \\ \midrule
golf              & -        & -             & 37.1          & \textbf{51.3} & -        & -             & -        & \textbf{50.5} \\ \midrule
kick              & -        & -             & 36.8          & \textbf{50.6} & -        & -             & -        & \textbf{48.3} \\ \midrule
lift              & -        & -             & \textbf{68.8} & 55.0          & -        & -             & -        & \textbf{51.4} \\ \midrule
ride              & 62.2     & \textbf{68.1} & 21.9          & 29.5          & 63.1     & \textbf{64.0} & -        & 30.6          \\ \midrule
run               & 50.2     & \textbf{61.4} & 20.1          & 34.3          & 48.1     & \textbf{61.9} & -        & 33.1          \\ \midrule
skate             & -        & -             & 13.0          & \textbf{40.0} & -        & -             & -        & \textbf{38.5} \\ \midrule
swing-b           & -        & -             & 32.7          & \textbf{54.8} & -        & -             & -        & \textbf{54.3} \\ \midrule
swing-s           & -        & -             & 16.4          & \textbf{19.3} & -        & -             & -        & \textbf{20.6} \\ \midrule
walk              & -        & -             & 28.3          & \textbf{39.5} & -        & -             & -        & \textbf{39.0} \\ \midrule
\textbf{Avg.}     & -        & -             & 31.8          & \textbf{42.1} & -        & -             & -        & \textbf{41.0} \\ \bottomrule
\end{tabular}
\caption{Results for \citet{ma2013action}. Action localization results measured as average IOU on UCF Sports dataset.}
\end{table}

\subsection{Spatiotemporal Deformable Part Models for Action Detection}

\citet{tian2013spatiotemporal} extend the concept of deformable parts model from 2D to 3D similar to \citet{ma2013action} but with some differences. The main difference is that this approach searches for a 3D subvolume considering parts both in space and time. SDPM also includes an explicit model to capture intra-class variation as a deformable configuration of parts. Finally, this approach shows effective resuls on action detection within a DPM framework without resorting to global BoW information, trajectories, or video segmentation.

The primary problem of generalizing DPM to 3D is that an action in a video may move spatially as frames progress. This is not a difficult problem in 2D as a static bounding box would cover most of the action parts. However, in videos, actions may move and a static learned bounding box will fail to cover the action across time. A naive approach would be to encapsulate the action with a large spatiotemporal box. However, that would drastically decrease the IOU of the prediction. The secondary problem is the difference between space and time. As the authors rightly point out, if an action size changes due to distance from camera, that does not mean the duration of the action changes as well. Thus, their feature pyramids employ multiple levels in space but not in time. Finally, they employ HOG3D feature d\cite{klaser2010learning} for their effectiveness. The HOG3D descriptors are based on a histogram of oriented spatiotemporal gradients as a volumetric generalization of the HOG \cite{dalal2005histograms} descriptor.

\begin{figure*}[t]
    \centering
    \includegraphics[width=0.90\linewidth]{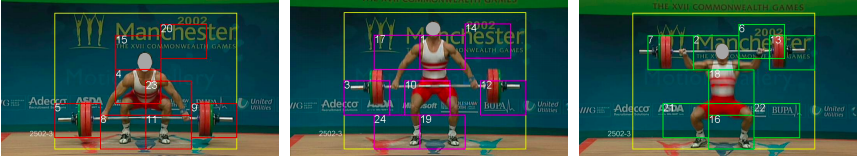}
    \caption{SDPM for \textbf{Lifting} in UCF-Sports dataset with parts learned in each temporal stage. There are total of 24 parts in this SDPM.}
    \label{fig:6-2}
\end{figure*}

\subsubsection{Root filter}
Following the DPM paradigm, the authors select a single bounding box for each video enclosing one cycle of given action. Volumes of other actions are treated as negative examples. Random volumes drawn from different scales of video are also added to negative samples to better discriminate action from background. The root filter is responsible for capturing the overall information of the action cycle by applying a SVM on the HOG3D features. An important aspect is to decide how to divide an action volume. Too few cells will decrease the overall discriminative power of the features while too many cells will prevent each cell from containing enough information to make it useful. The size of the spatial dimension in the root filter can be determined empirically. The authors used a 3x3xT size. This cannot be done for temporal dimension as an action may vary from a 5-30 seconds. Thus, the size of this filter must be determined automatically depending on the distribution of the action in the training set.

\subsubsection{Deformable parts}
The authors observed that extracting HOG3D features from part models at twice the resolution and with more cells in space enabled the learned parts to capture important details. A point to note is that parts selected by this model are \textbf{allowed} to overlap in space. After SVM, subvolumes with higher weights (more discriminative power for a given action type) are selected as parts. The authors divided action cells into 12x12xT cells to extract HOG3D features and each part occupies 3x3x1 cell. Then, they greedily selected N parts with highest weights that fill 50\% of action cycle volume. Parts weights initialization is corresponding to the cell weights containted inside them. An anchor position $(x_i, y_i, t_i)$ for $i$-th part is also determined. To address intra-class variation, the authors use a quadratic function to allow parts to shift within a certain spatiotemporal region.

\subsubsection{Action detection with SDPM}
Given a test video, SDPM builds a spatiotemporal feature pyramid by computing HOG3D features at different scales. Template matching during detection is done using a sliding window approach. Score maps for root and part filters are computed at every level of the pyramid using template matching. For level $l$, the score map $S(l)$ of each filter can be obtained by correlation of filter $F$ with features of test video volume $\phi(l)$,

\begin{equation}
    S(l, i, j, k) = \sum_{m, n, p} F(i, j, k) \phi(i+m, j+n, k+p, l)
\end{equation}

At level $l$ in feature pyramid, score of detection volume centered at $(x, y, t)$ is sum of the score of root filter on this volume and scores from each part filter on best possible subvolume:

\begin{equation}
    \begin{aligned}
    score(x, y, t, l) = F_{0} \cdot \alpha(x, y, t, l) + \\
    \sum_{i \le i \le n} \max_{(x', y', t') \in Z} [F_{i} \cdot \beta(x'_{i}, y'_{i}, t'_{i}, l) - \varepsilon (i, X_{i})]
    \end{aligned}
\end{equation}
where $F_{0}$ is the root filter and $F_{i}$ are part filters. $\alpha(x, y, t, l)$ and $\beta(x, y, t, l)$ are features of a 3x3xT volume centered at $(x, y, t)$ and 3x3x1 volume centered at part location $(x', y', t')$ respectively, at level $l$ of feature pyramid. $Z$ is the set of all different possible part locations and $\varepsilon (i, X_{i})$ is corresponding deformation cost. Highest score is chosen at the end based on a threshold. A scanning search algorithm is employed instead of exhaustive search.

\subsubsection{Results}

The authors present their results on Weizmann \cite{ActionsAsSpaceTimeShapes_iccv05}, UCF Sports \cite{rodriguez2008action}, and MSR-II \cite{zhang2016large} datasets. Without much surprise, SDPM achieves 100\% accuracy on the Weizmann dataset as the challenge is easy (9 actions on static background). On the UCF-Sports dataset, the authors achieved an average classification accuracy of \textbf{75.2\%} which is higher than \citet{ma2013action} (73.1\%) but lower than \citet{raptis2012discovering} (79.4\%). On the MSR-II dataset, they outperformed model without parts as well as baselines.



\section{Graph-Based Models}

\subsection{Action localization in videos through context walk}

\citet{soomro2015action} take a different approach to action localization. As a brief summary, they over-segment videos into supervoxels, learn context relationships (background-background and background-foreground), estimate probability of supervoxel belonging to an action for each supervoxel to create a conditional distribution of an action over all supervoxels, use a \textbf{Conditional Random Field} (CRF) to find action proposals in video, and use a \textbf{SVM} to obtain confidence scores. This \textit{context walk} eliminates the need to use a \textit{sliding window} approach and do an exhaustive search over an entire video. This is useful because most videos have under 20\% of frames with actions in them.

\subsubsection{Context Graphs for Training Videos}
Assuming index of training videos for action $c = 1...C$ is between range $n = 1...N_{c}$, where $N_{c}$ is number of training videos for action $c$, the $i$-th supervoxel in the $n$-th video is represented by $\textbf{u}_{n}^{i}, i = 1...I_{n}$, where $I_{n}$ is number of supervoxels in video $n$. Each supervoxel either belongs to foreground action or background. The authors now construct a directed graph $\textbf{G}_{n}(\textbf{V}_{n}, \textbf{E}_{n})$ for each training video across all action classes. Nodes in the graph are represnted by supervoxels while edges $\textbf{e}^{ij}$ emanate from all nodes belonging to foreground.

Let each supervoxel \textbf{u} be represented by its spatiotemporal centroid, $\textbf{u}_{n}^{i} = (x_{n}^{i}, y_{n}^{i}, t_{n}^{i})$. The features associated with $\textbf{u}_{n}^{i}$ are given by $\mathbf{\Phi}_{n}^{i} = ( _{1}\phi_{n}^{i}, _{2}\phi_{n}^{i},...,_{F}\phi_{n}^{i})$, where $F$ is total number of features. Graphs $\textbf{G}_{n}$ and $\mathbf{\Phi}_{n}^{i} \forall n=1...N_{c}$ are represented by composite graph $H_{c}$ which contains all information necessary for action localization.

\subsubsection{Context Walk in Testing Video}
The model obtains about ~200-300 supervoxels per video. The goal is to visit each supervoxel in sequence, referred to as a \textit{context walk}. The initial supervoxel is selected randomly and similar supervoxels are found by nearest neighbor algorithm. The following function $\psi(\cdot)$ generates a conditional distribution over all supervoxels in testing video given only current supervoxel $\mathbf{v}^{\tau}$, features $\Phi^{\tau}$, and composite graph $\mathbf{H}$:

\begin{equation}
    \begin{aligned}
    \psi(\mathbf{v}|\mathbf{v}^{\tau}, \mathbf{\Phi}^{\tau}, \mathbf{H}, \mathbf{w}_{\psi}) = \\
    Z^{-1} \sum_{n=1}^{N_{c}} \sum_{i=1}^{I_{n}} \sum_{j|e_{ij} \in \mathbf{E}_{n}} H_{\sigma}(\mathbf{\Phi}^{\tau}, \mathbf{\Phi}_{n}^{i}; \mathbf{w}_{\sigma}) \\
    \cdot H_{\sigma}(\mathbf{v}, \mathbf{v}^{\tau}, \mathbf{u}_{n}^{i}, \mathbf{u}_{n}^{j}; w_{\delta})
    \end{aligned}
\end{equation}

where $H_{\sigma}$ computes similarity between features of current supervoxel in testing video. Skipping ahead to inference, the supervoxel with highest probability is selected in next step of context walk:

\begin{equation}
    \mathbf{v}^{\tau + 1} = \argmax_{\mathbf{v}} \Psi^{\tau}(v|\mathbf{S}_{\mathbf{v}}^{\tau}, \mathbf{S}_{\mathbf{\Phi}}^{\tau}, \mathbf{H}, \mathbf{w})
\end{equation}

\subsubsection{Measuring Supervoxel Action Specificity}
The authors quantify discriminative supervoxels based on an action specificity score. If $\xi(k_c)$ represents the ratio of number of supervoxels from foreground of action $c$ in cluster $k_c$ to all supervoxels from action $c$ in cluster, then, given appearance/motion descriptors \textbf{d}, if supervoxel belongs to cluster $k_c$, its action specificity $H_{\chi}(\mathbf{v}^i)$ is quantified as:

\begin{equation}
    H_{\chi}(\mathbf{v}^i) = \xi(k_c) \cdot \exp(\frac{\left \| \mathbf{d}^i - \mathbf{d}_{k_c} \right \|}{r_{k_c}})
\end{equation}
where $\textbf{d}_{k_c}$ and $r_{k_c}$ are center and radius of $k$-th cluster, respectively.

\subsubsection{Inferring Action Locations using 3D-CRF}
Once we have conditional distribution $\mathbf{\Psi^T(\cdot)}$, we can merge supervoxels belonging to actions to create a continuous flow of supervoxels without any gaps. The authors use CRFs for this purpose. They minimize the negative log likelihood over all supervoxel labels \textbf{a} in the video:

\begin{equation}
    \begin{aligned}
    -\log(Pr(\mathbf{a}|\mathbf{G}, \mathbf{\Phi}, \mathbf{\Psi}^T; w_\gamma )) = \sum_{\mathbf{v}^i \in \mathbf{V}} (\Theta(a^i|\mathbf{v}^i, \mathbf{\Psi}^T) + \\
    \sum_{v^j|e^{ij} \in \mathbf{E}} \gamma(a^i, a^j | \mathbf{v}^i, \mathbf{v}^j, \mathbf{\Phi}^i, \mathbf{\Phi}^j; w_\gamma))
    \end{aligned}
\end{equation}
where $\Theta(\cdot)$ captures unary potential and depends on conditional distribution after $T$ steps and action specificity measured above. Both are normalized between 0 and 1.

\begin{figure*}[t]
    \centering
    \includegraphics[width=0.90\linewidth]{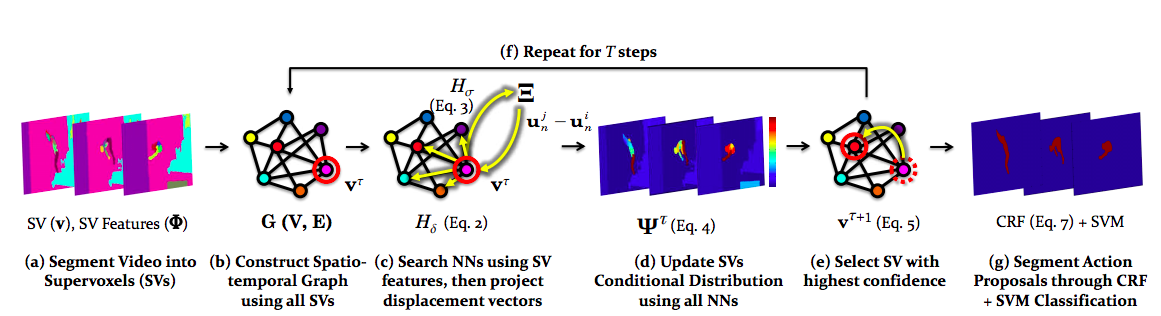}
    \caption{Pipeline for method proposed by \citet{soomro2015action}.}
    \label{fig:7-1}
\end{figure*}

\subsubsection{Results}
The approach is evaluated on UCF-Sports \cite{rodriguez2008action}, sub-JHMDB \cite{jhuang2013towards}, and THUMOS'13 \cite{jiang2014thumos} datasets. The biggest advantage of this method is the complexity. While SDPM by \citet{tian2013spatiotemporal} and Tubelets by \citet{jain2014action} have complexities $\mathcal{O}(n^4)$ and $\mathcal{O}(n^2)$ respectively, this work has complexity $\mathcal{O}(c)$ where $c$ is the number of classifier evaluations.

\begin{table}[ht]
\centering
\small
\begin{tabular}{|c|c|c|}
\hline
Method                & UCF-Sports & sub-JHMDB \\ \hline
\citet{wang2014video}           & 47\%       & 36\%      \\ \hline
\citet{wang2014video} (iDTF+FV) & -          & 34\%      \\ \hline
\citet{jain2014action}           & 53\%       & -         \\ \hline
\citet{tian2013spatiotemporal}           & 42\%       & -         \\ \hline
\citet{lan2011discriminative}            & 38\%       & -         \\ \hline
\textbf{\citet{soomro2015action}}               & \textbf{55}\%       & \textbf{42}\%      \\ \hline
\end{tabular}
\caption{\citet{soomro2015action}. Comparison of methods at 20\% overlap.}
\label{Table:7-1}
\end{table}

\subsection{Spatial Temporal Graph Convolutional Networks for Skeleton-Based Action Recognition}

Besides optical flow and traditional pixel level information, there is a class of representations based on human skeleton and joints. These form conceptual graphs that can be used to classify actions. This paper by \citet{yan2018spatial} uses those features to classify and localize actions.

\textbf{Graph neural networks} are a recent paradigm that generalize convolutional neural networks to graphs of arbitrary structures. They have shown to perform well on tasks such as image classification, document classification, and semi-supervised learning. \cite{yan2018spatial} extend the idea of graph neural networks to \textit{Spatial-Temporal Graph Convolutional Networks (ST-GCN)} which attempt to model, localize, and classify actions. The graph representation contains \textit{spatial edges} that conform to natural connectivity of joints and \textit{temporal edges} that connect to same joints across consecutive time steps. Besides ST-GCN, the authors also introduce several principles to design convolution kernels in ST-GCN. Finally, they evaluate their models on large scale datasets to demonstrate the approach's effectiveness as we shall see.

\subsubsection{Spatial Temporal Graph ConvNet}
The overall pipeline expects skeleton based data obtained from a motion-capture device or pose estimation algorithm from videos. For each frame, there will be a set of join coordinates. Given these sequences of body joints, the model constructs a spatial temporal graph with joints as graph nodes and natural connectivities in both human body structures and time as graph edges.

\subsubsection{Skeleton Graph Construction}
The authors create an undirected spatial temporal graph $G = (V, E)$ on a skeleton sequence with $N$ joints and $T$ frames feature both intra-body and inter-frame connection. The node set $V = {v_{ti}|t=1,...,T,i=1,...,N}$ includes all joints in skeleton sequence. As ST-GCN's input, feature vector on node $F(v_{ti})$ consists of coordinate vectors as well as estimation confidence on $i$-th join on frame $t$. The construction of the graph is divided into two steps: The joints within one frame are connected with edges according to human body structure. Then, each joint is connected to the same joint in the consecutive time step's graph. Connections are naturally made without manual intervention. This also provides generalization capabilities with respect to different datasets. Formally, the edge set $E$ is composed of two subsets: $E_{S} = {v_{ti}v{tj}|(i, j) \in H}$ consisting of intra-skeleton connection at each frame where $H$ is set of naturally connected human body joints. The second subset consists of inter-frame edges connecting same joints in consecutive frames and is expressed as $E_F = {v_{ti}v_{(t+1)i}}$.  

\subsubsection{Spatial Graph Convolutional Neural Network}
Let us just consider graph CNN model within one single frame. At a single frame at time $\tau$, there will be $N$ joint nodes $V_t$ along with skeleton edges $E_S(\tau) = {v_{ti}v{tj}|(i, j) \in H}$. Given a convolution operator with kernel size $K \times K$, and input feature map $f_{in}$ with number of channels c, the output value of single channel at spatial location $x$ can be written as:

\begin{equation}
    f_{out}(\mathbf{x}) = \sum_{h=1}^{K}\sum_{w=1}^{K}f_{in}(\mathbf{p}(\mathbf{x}, h, w) \cdot \mathbf{w}(h, w))
\end{equation}
where \textbf{sampling function} $\mathbf{p}:Z^2 \times Z^2 \rightarrow Z^2$ enumerates neighbors of location $x$. The \textbf{weight function} $\mathbf{w}: Z^2 \rightarrow \mathbb{R}^c$ provides weight vector in $c$-dimensional real space for computing inner product with sampled input feature vector of dimension $c$. Standard convolution on image domain is achieved by encoding a rectangular grid in \textbf{p(x)}. Please refer to the original paper for reformulation of sampling and weight functions on 2D image domains. Now, we can write a graph convolution as:

\begin{equation}
    f_{out}(v_{ti}) = \sum_{v_{tj} \in B(v_{ti})} \frac{1}{Z_{ti}(v_{tj})} f_{in}(v_{tj}) \cdot \mathbf{w}(l_{ti}(v_{tj}))
\end{equation}
where normalizing term $Z_{ti}(v_{tj}) =| {v_{tk}|l_{ti}(v_{tk}) = l_{ti}(v_{tj})}$ equals cardinality of corresponding subset. To model the temporal aspect of this graph, we simply use the same sampling function and labeling map $l_{ST}$. Because temporal axis is well-ordered, we directly modify label map for spatial temporal neighborhood rooted at $v_{ti}$ to be:

\begin{equation}
    l_{ST}(v_{qj}) = l_{ti}(v_{tj})+(q-t+ \floor{\Upgamma / 2}) \times K
\end{equation}
where $l_{ti}(v_{tj})$ is label map for single frame case at $v_{ti}$.

\subsubsection{Implementing ST-GCN}
The implementation of graph convolution is the same as in \citet{kipf2016semi}. The intra-body connections are represented by an adjacency matrix \textbf{A} and identity matrix \textbf{I}. Thus, in the single frame case,

\begin{equation}
    f_{out} = \mathbf{\Lambda}^{-\frac{1}{2}}(\mathbf{A} + \mathbf{I}) \mathbf{\Lambda}^{-\frac{1}{2}}\mathbf{f}_{in}\mathbf{W}
\end{equation}
where $\Lambda^{ii} = \sum_{j}(A^{ij}+I^{ij})$. 

In the multiple subset case,

\begin{equation}
    f_{out} = \sum_{j} \mathbf{\Lambda}_{j}^{-\frac{1}{2}}\mathbf{\Lambda}_{j}\mathbf{\Lambda}_{j}^{-\frac{1}{2}}f_{in}\mathbf{W}_j
\end{equation}
where similarity $\Lambda_{j}^{ii} = \sum_{k}(A_{j}^{jk}) + \alpha$. Here, the authors set $\alpha = 0.001$ to avoid empty rows in $\mathbf{A}_j$.

The input is first fed to a batch normalization layer to normalize data. The ST-GCN model is composed of 9 layers of spatial temporal graph convolution operations. The first three layers have 64 output channels, second three layers have 128 channels, and last three layers have 256 output channels. These layers have 9 temporal kernel size. The ResNet mechanism is applied to each of these layers. There is also a random dropout of 0.5 after each layer to prevent overfitting. The 4-th and 7-th layer have strides 2 for pooling. Then, a global pooling is performed to achieve a 256 dimension feature tensor for each sequence. Lastly, the feature vector is fed to a Softmax layer for classification. The model is optimized using stochastic gradient descent with a learning rate of 0.01 and decayed by 0.1 after every 10 epochs.

\subsubsection{Results}
The authors tested the model on \textbf{Kinetics human action dataset} \cite{kay2017kinetics} and \textbf{NTU-RGB+D} \cite{shahroudy2016ntu} dataset. On the Kinetics dataset, ST-GCN achieved a 10.4\% and 12.8\% increase in Top-1 and Top-5 accuracies when compared to frame based methods. On the NTU-RGB+D dataset, they achieved a 1.9\% and 3.5\% increase on X-Sub and X-View accuracies when compared to all previous methods. 

\begin{table}[ht]
\centering
\small
\begin{tabular}{|c|c|c|}
\hline
               & Top-1         & Top-5         \\ \hline
RGB \citet{kay2017kinetics}           & 57.0          & 77.3          \\ \hline
Optical Flow \citet{kay2017kinetics}   & 49.5          & 71.9          \\ \hline
Feature Enc. \citet{fernando2015modeling}   & 14.9          & 25.8          \\ \hline
Deep LSTM \citet{liu2016spatio}     & 16.4          & 35.3          \\ \hline
Temporal Conv. \citet{kim2017interpretable} & 20.3          & 40.0          \\ \hline
ST-GCN         & \textbf{30.7} & \textbf{52.8} \\ \hline
\end{tabular}
\caption{Results for \citet{yan2018spatial}Action recognition performance on skeleton based models on Kinetics dataset. The first two methods are frame based methods.}
\end{table}

\section{3D Convolutional Neural Networks}



\subsection{A Closer Look at Spatiotemporal Convolutions for Action Recognition}

Let us look into an approach which completely uses convolutional neural networks without any special feature representations. The paper by \citet{tran2017closer} introduces an even more advanced approach to action recognition with a demonstration of a new form of convolution. The method is aimed for action recognition only. However, it can be supplemented with additional features to enable \textit{action localization}.

The authors mainly focus on the domain of residual learning for action recognition. They explore the existing types of 3D convolutions and namely introduce two new types of convolution. The first new convolution is a mixed convolution where early layers of the model perform 3D convolutions while later layers perform spatial or 2D convolutions over the learned features. This is called the \textit{MC} or mixed convolution. The second new convolution is a complete decomposition of the 3D convolution into separate 2D spatial convolution and 1D temporal convolution. This is called the \textit{R(2+1)D} convolution. This decomposition brings in two advantages. Firstly, the decomposition introduces an additional nonlinear rectification between two operations. This means, you double the number of nonlinearities compared to a network using full 3D convolutions for same number of parameters. Secondly, this facilitates optimization leading to lower training loss and lower testing loss. Let us now explore the various types of convolutions for videos.

\subsubsection{Convolutional residual blocks for video}
Within the framework of residual learning, there are several spatiotemporal convolution variants available. Let \textbf{x} denote input clip of size $3 \times L \times H \times W$, where $L$ is number of frames in clip, $H$ and $W$ are frame height and width, and 3 refers to the RGB channels. Let $\textbf{z}_i$ be tensor computed by $i$-th convolutional block. Then, the output of that block is:

\begin{equation}
    \mathbf{z}_i = \mathbf{z}_{i-1} + \mathcal{F}(\mathbf{z}_{i-1}; \theta_i)
\end{equation}
where $\mathcal{F}(;\theta_i)$ implements composition of two convolutions parameterized by weights $\theta_{i}$ and application of ReLU functions.

\textbf{R2D: 2D convolutions over the entire clip. }
2D CNNs for video ignore temporal ordering and treat $L$ frames independently of channels. This is basically reshaping the input 4D tensor \textbf{x} into a 3D tensor of size $3L \times H \times W$. The output $\mathbf{z}_i$ of $i$-th block is also a 3D tensor. Each filter is 3D and has size $N_{i-1} \times d \times d$, where d denotes spatial width and height. Even tho filter is 3D, it only convolves in 2D over the \textit{spatial} dimensions. All temporal information of video is collapsed into single-channel feature maps. This prevents any sort of temporal reasoning.

\textbf{f-R2D: 2D convolutions over frames. }
Another 2D CNN approach involves processing independently the $L$ frames via a series of 2D convolutional residual block. Same filterers are applied to all $L$ frames. No temporal modeling is performed on the convolutional layers and global spatiotemporal pooling layer at the end simply fuses information extracted independently from the $L$ frames. This architecture variant is referred to as f-R2D (frame-based R2D).

\textbf{R3D: 3D convolutions. }
3D CNNs \cite{tran2015learning} preserve temporal information and propagate it through the layers of the network. The tensor $\mathbf{z}_i$ is 4D in shape and has size $N_i \times L \times H_i \times W_i$, where $N_i$ is number of filters used in $i$-th block. Each filter is 4-dimensional and has size $N_{i-1} \times t \times d \times d$ where $t$ denotes the temporal extent of the filter (the authors used $t = 3$).

\textbf{M$C_x$ and rM$C_x$: mixed 3D-2D convolutions}
The intuition behind \textbf{MC} layers is that in early layers, motion modeling may be important while in later layers, motion or temporal modeling is not necessary. In the authors' experiments of a 5 layer residual block, two variants come out where first three layers are 3D convolutions while last two are 2D. The other variant is just the opposite of it.

\subsubsection{R(2+1)D: (2+1)D convolutions}
Another hypothesis proposed by the authors is that full 3D convolutions can be more conveniently approximated by a 2D convolution followed by a 1D convolution. Thus, they designed a R(2+1)D architecture where the $N_i$ 3D convolutional filter of size $N_{i-1} \times t \times d \times d$ is replaced with a (2+1)D block consisting of $M_i$ 2D convolutional filters of size $N_{i-1} \times 1 \times d \times d$ and $N_i$ temporal convolutional filters of size $M_i \times t \times 1 \times 1$. The hyperparameter $M_i$ determines dimensionality of intermediate sub-space where signal is projefcted between spatial and temporal convolutions. The authors choose $M_i = \floor{\frac{td^2N_{i-1}N{i}}{d^2N_{i-1}+tN_{i}}}$ so that number of parametrs in the block approximately equal to the 3D variant. This spatiotemporal decomposition can be applied to any 3D convolutional layer.

\begin{figure*}[t]
    \centering
    \includegraphics[width=0.8\linewidth]{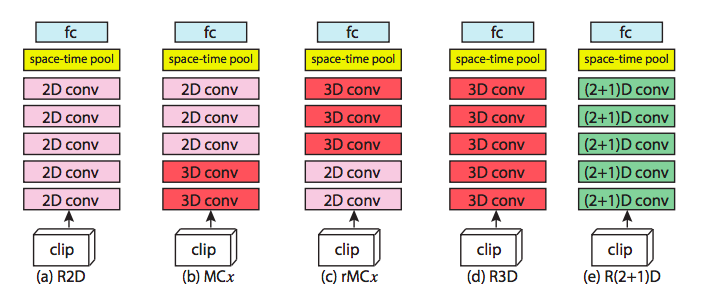}
    \caption{Residual network architectures for video classification. (a) R2D are 2D ResNets; (b) MCx are ResNets with mixed convolutions; (c) rMCx use reversed mixed convolutions; (d) R3D are 3D ResNets; and (e) R(2+1)D are ResNets with (2+1)D convolutions.}
    \label{fig:8-1}
\end{figure*}

\subsubsection{Results}
The authors experimented their new architecture on the Kinetics \cite{kay2017kinetics} and Sports-1M \cite{karpathy2014large} datasets. They also pre-trained the models on these two datasets and then finetuned them to UCF-101 \cite{soomro2012ucf101} and HMDB51 \cite{kuehne2013hmdb51} datasets.

The networks experimented with are the ResNet-18 and ResNet-34 network architectures. Frame input size is $112 \times 112$. The authors use one spatial downsampling of $1 \times 2 \times 2$, and three spatiotemporal downsampling with convolutional striding of $2 \times 2 \times 2$. For training, $L$ consecutive frames are randomly sampled. Batch normalization is applied to all convolutional layers and batch size is set to 32 per GPU. The initial learning rate is set to 0.01 and is decayed by 0.1 every 10 epochs. The R(2+1)D layer architecture reported an average 3\% improvement from previous methods. 


\section{Action Localization Datasets}
Let us explore common datasets used for action localization. This can help us understand what sort of features and data is available to explore algorithms and models upon. We shall explore seven datasets that are of different sizes, domains, and contain different features.

\subsubsection{Kinetics Dataset}
The first dataset we explore is the Kinetics dataset by \citet{kay2017kinetics}. The dataset is sourced from YouTube videos to encourage variation between videos in same and different action classes. There are \textbf{400} actions, minimum 400 video clips per action, and contains 306,245 videos in total. The way the dataset was built was through first curating action classes by merging different previous dataset action classes. Secondly, the videos were sourced from YouTube corpus. To collect the best videos, an aggregation of relevance feedback scores were used with multiple queries. Finally, human tagging was used to manually annotate the videos for accuracy and consistent. 

\subsubsection{Weizzman Dataset}
The Weizzman dataset by \citet{ActionsAsSpaceTimeShapes_iccv05} contains \textbf{90} low-resolution video sequences $(180 \times 144, 50 fps)$ consisting of 10 different action classes. Some actions are: “running”, “walking”, “jumpingjack”,
“jumping-forward-on-two-legs”, “jumping-in-placeon-two-legs”,
“galloping-sideways”, “waving-two-hands”,
“waving-one-hand”, “bending”.

\subsubsection{UCF-101 Dataset}
The UCF-101 by \citet{soomro2012ucf101} is arguably one of the most famous datasets for action recognition and action localization. As its name implies, it consists of \textbf{101} action classes, 13,320 clips, 4-7 clips per group, a mean clip length of 7.21 seconds, total duration of 1600 minutes, frame rate of 25 fps, and resolution of $320 \times 240$. The source of the videos is the YouTube corpus. It is an extension of the UCF-50 dataset.

\subsubsection{UCF-Sports Dataset}
The UCF-Sports dataset by \citet{rodriguez2008action} is an video dataset containing \textbf{10} actions primarily in the sports domain. There are 150 clips, mean clip length of 6.39s, frame rate of 10fps, total duration of 958s, and at a resolution of $720 \times 480$. The maximum and minimum number of clips per class are 22 and 6 respectively. The videos are sourced through BBC and ESPN video corpus.

\subsubsection{THUMOS'14 Dataset}
\citet{jiang2014thumos} released the THUMOS'14 dataset which contains \textbf{101} actions, 13,000 temporally trimmed videos, over 1000 temporally untrimmed videos, over 2500 negative sample videos, and bounding boxes for 24 action classes.

\subsubsection{HMDB Dataset}
\citet{jhuang2013towards} introduce the HMDB dataset containing \textbf{51} actions, 6849 clips, and each class contains at least 101 clips. Actions include laughing, talking, eating drinking, pull up, sit down, ride bike, etc.

\subsubsection{Activity Net Dataset}
\citet{caba2015activitynet} curated the Activity Net dataset which contains \textbf{200} action classes, 100 untrimmed videos per class, 1.54 activity instances per video on average, and a total of 38,880 minutes of videos. This dataset was hosted as a challenge at CVPR 2018.

\subsubsection{NTURGB-D Dataset}
Finally, we look at the NTURGB-D dataset by \citet{shahroudy2016ntu}. The dataset contains 56,880 action samples distributed across \textbf{60} actions with each video containing the following data:
\begin{enumerate}
    \item RGB videos
    \item depth map sequences
    \item 3D skeletal data
    \item infrared videos
\end{enumerate}
Video samples are of resolution $1920 \times 1080$, depth map and IR videos have resolution $512 \times 424$, and 3D skeletal data have three dimensional locations of 25 major body parts.


\section{Conclusion}

In conclusion, we explored eight approaches to action localization. There are many more methods and techniques that are used to solve this problem. However, majority of them rely upon clever usage of the same types of features including RGB pixel values, optical flow, skeleton graphs, etc. Action proposal networks are effective but they are expensive and usually need an exhaustive search of the entire video. If the video length is large (e.g. CCTV camera footage spanning over several hours), these approaches turn out to be computationally infeasible. Figure centric models try to solve this problem. However, they require too much of manual feature construction to be automated on a large scale. Deformable parts models actually solve the problem of selective sampling and extracting segments for action localization. There is room for improvement there too. Graph based models also lower the amount of search and classification operations with optimizations. However, they require skeletal data which means a pose estimation algorithm is required as a pre-processing step. The accuracy of the pose estimation algorithm can also create bias towards the training and inference of the model. Finally, spatiotemporal convolutions provide an interesting proposition and possibly we can extend this technique while incorporating more features to solve the problem of action localization. 

{\small
\bibliography{egbib}

\begin{thebibliography}{46}
\providecommand{\natexlab}[1]{#1}
\providecommand{\url}[1]{#1}
\csname url@samestyle\endcsname
\providecommand{\newblock}{\relax}
\providecommand{\bibinfo}[2]{#2}
\providecommand{\BIBentrySTDinterwordspacing}{\spaceskip=0pt\relax}
\providecommand{\BIBentryALTinterwordstretchfactor}{4}
\providecommand{\BIBentryALTinterwordspacing}{\spaceskip=\fontdimen2\font plus
\BIBentryALTinterwordstretchfactor\fontdimen3\font minus
  \fontdimen4\font\relax}
\providecommand{\BIBforeignlanguage}[2]{{%
\expandafter\ifx\csname l@#1\endcsname\relax
\typeout{** WARNING: IEEEtranN.bst: No hyphenation pattern has been}%
\typeout{** loaded for the language `#1'. Using the pattern for}%
\typeout{** the default language instead.}%
\else
\language=\csname l@#1\endcsname
\fi
#2}}
\providecommand{\BIBdecl}{\relax}
\BIBdecl

\bibitem[Tian et~al.(2013)Tian, Sukthankar, and Shah]{tian2013spatiotemporal}
Y.~Tian, R.~Sukthankar, and M.~Shah, ``Spatiotemporal deformable part models
  for action detection,'' in \emph{Computer Vision and Pattern Recognition
  (CVPR), 2013 IEEE Conference on}.\hskip 1em plus 0.5em minus 0.4em\relax
  IEEE, 2013, pp. 2642--2649.

\bibitem[Ma et~al.(2013)Ma, Zhang, Ikizler-Cinbis, and Sclaroff]{ma2013action}
S.~Ma, J.~Zhang, N.~Ikizler-Cinbis, and S.~Sclaroff, ``Action recognition and
  localization by hierarchical space-time segments,'' in \emph{Computer Vision
  (ICCV), 2013 IEEE International Conference on}.\hskip 1em plus 0.5em minus
  0.4em\relax IEEE, 2013, pp. 2744--2751.

\bibitem[Ke et~al.(2007)Ke, Sukthankar, and Hebert]{ke2007event}
Y.~Ke, R.~Sukthankar, and M.~Hebert, ``Event detection in crowded videos,'' in
  \emph{Computer Vision, 2007. ICCV 2007. IEEE 11th International Conference
  on}.\hskip 1em plus 0.5em minus 0.4em\relax IEEE, 2007, pp. 1--8.

\bibitem[Kl{\"a}ser et~al.(2010)Kl{\"a}ser, Marsza{\l}ek, Schmid, and
  Zisserman]{klaser2010human}
A.~Kl{\"a}ser, M.~Marsza{\l}ek, C.~Schmid, and A.~Zisserman, ``Human focused
  action localization in video,'' in \emph{European Conference on Computer
  Vision}.\hskip 1em plus 0.5em minus 0.4em\relax Springer, 2010, pp. 219--233.

\bibitem[Lan et~al.(2011)Lan, Wang, and Mori]{lan2011discriminative}
T.~Lan, Y.~Wang, and G.~Mori, ``Discriminative figure-centric models for joint
  action localization and recognition,'' in \emph{Computer Vision (ICCV), 2011
  IEEE International Conference on}.\hskip 1em plus 0.5em minus 0.4em\relax
  IEEE, 2011, pp. 2003--2010.

\bibitem[Prest et~al.(2013)Prest, Ferrari, and Schmid]{prest2013explicit}
A.~Prest, V.~Ferrari, and C.~Schmid, ``Explicit modeling of human-object
  interactions in realistic videos,'' \emph{IEEE transactions on pattern
  analysis and machine intelligence}, vol.~35, no.~4, pp. 835--848, 2013.

\bibitem[Gkioxari and Malik(2015)]{gkioxari2015finding}
G.~Gkioxari and J.~Malik, ``Finding action tubes,'' in \emph{Computer Vision
  and Pattern Recognition (CVPR), 2015 IEEE Conference on}.\hskip 1em plus
  0.5em minus 0.4em\relax IEEE, 2015, pp. 759--768.

\bibitem[Yu and Yuan(2015)]{yu2015fast}
G.~Yu and J.~Yuan, ``Fast action proposals for human action detection and
  search,'' in \emph{Proceedings of the IEEE conference on computer vision and
  pattern recognition}, 2015, pp. 1302--1311.

\bibitem[Jain et~al.(2014)Jain, Van~Gemert, J{\'e}gou, Bouthemy, and
  Snoek]{jain2014action}
M.~Jain, J.~Van~Gemert, H.~J{\'e}gou, P.~Bouthemy, and C.~Snoek, ``Action
  localization with tubelets from motion,'' in \emph{CVPR-International
  Conference on Computer Vision and Pattern Recognition}, 2014.

\bibitem[Weinzaepfel et~al.(2015)Weinzaepfel, Harchaoui, and
  Schmid]{weinzaepfel2015learning}
P.~Weinzaepfel, Z.~Harchaoui, and C.~Schmid, ``Learning to track for
  spatio-temporal action localization,'' in \emph{Proceedings of the IEEE
  international conference on computer vision}, 2015, pp. 3164--3172.

\bibitem[Yan et~al.(2018)Yan, Xiong, and Lin]{yan2018spatial}
S.~Yan, Y.~Xiong, and D.~Lin, ``Spatial temporal graph convolutional networks
  for skeleton-based action recognition,'' \emph{arXiv preprint
  arXiv:1801.07455}, 2018.

\bibitem[Soomro et~al.(2015)Soomro, Idrees, and Shah]{soomro2015action}
K.~Soomro, H.~Idrees, and M.~Shah, ``Action localization in videos through
  context walk,'' in \emph{Computer Vision (ICCV), 2015 IEEE International
  Conference on}.\hskip 1em plus 0.5em minus 0.4em\relax IEEE, 2015, pp.
  3280--3288.

\bibitem[Diba et~al.(2017)Diba, Sharma, and Van~Gool]{diba2017deep}
A.~Diba, V.~Sharma, and L.~Van~Gool, ``Deep temporal linear encoding
  networks,'' in \emph{Computer Vision and Pattern Recognition}, 2017.

\bibitem[Ren et~al.(2015)Ren, He, Girshick, and Sun]{ren2015faster}
S.~Ren, K.~He, R.~Girshick, and J.~Sun, ``Faster r-cnn: Towards real-time
  object detection with region proposal networks,'' in \emph{Advances in neural
  information processing systems}, 2015, pp. 91--99.

\bibitem[Xu and Corso(2012)]{xu2012evaluation}
C.~Xu and J.~J. Corso, ``Evaluation of super-voxel methods for early video
  processing,'' in \emph{Computer Vision and Pattern Recognition (CVPR), 2012
  IEEE Conference on}.\hskip 1em plus 0.5em minus 0.4em\relax IEEE, 2012, pp.
  1202--1209.

\bibitem[Zhang et~al.(2016)Zhang, Li, Wang, Ogunbona, Liu, and
  Tang]{zhang2016large}
J.~Zhang, W.~Li, P.~Wang, P.~Ogunbona, S.~Liu, and C.~Tang, ``A large scale
  rgb-d dataset for action recognition,'' in \emph{Proc. ICPR}, 2016.

\bibitem[Hou et~al.(2017)Hou, Chen, and Shah]{hou2017tube}
R.~Hou, C.~Chen, and M.~Shah, ``Tube convolutional neural network (t-cnn) for
  action detection in videos,'' in \emph{IEEE International Conference on
  Computer Vision}, 2017.

\bibitem[Girshick et~al.(2014)Girshick, Donahue, Darrell, and
  Malik]{girshick2014rich}
R.~Girshick, J.~Donahue, T.~Darrell, and J.~Malik, ``Rich feature hierarchies
  for accurate object detection and semantic segmentation,'' in
  \emph{Proceedings of the IEEE conference on computer vision and pattern
  recognition}, 2014, pp. 580--587.

\bibitem[Tran et~al.(2015)Tran, Bourdev, Fergus, Torresani, and
  Paluri]{tran2015learning}
D.~Tran, L.~Bourdev, R.~Fergus, L.~Torresani, and M.~Paluri, ``Learning
  spatiotemporal features with 3d convolutional networks,'' in
  \emph{Proceedings of the IEEE international conference on computer vision},
  2015, pp. 4489--4497.

\bibitem[Rodriguez et~al.(2008)Rodriguez, Ahmed, and Shah]{rodriguez2008action}
M.~D. Rodriguez, J.~Ahmed, and M.~Shah, ``Action mach a spatio-temporal maximum
  average correlation height filter for action recognition,'' in \emph{Computer
  Vision and Pattern Recognition, 2008. CVPR 2008. IEEE Conference on}.\hskip
  1em plus 0.5em minus 0.4em\relax IEEE, 2008, pp. 1--8.

\bibitem[Jhuang et~al.(2013)Jhuang, Gall, Zuffi, Schmid, and
  Black]{jhuang2013towards}
H.~Jhuang, J.~Gall, S.~Zuffi, C.~Schmid, and M.~J. Black, ``Towards
  understanding action recognition,'' in \emph{Proceedings of the IEEE
  international conference on computer vision}, 2013, pp. 3192--3199.

\bibitem[Soomro et~al.(2012)Soomro, Zamir, and Shah]{soomro2012ucf101}
K.~Soomro, A.~R. Zamir, and M.~Shah, ``Ucf101: A dataset of 101 human actions
  classes from videos in the wild,'' \emph{arXiv preprint arXiv:1212.0402},
  2012.

\bibitem[Jiang et~al.(2014)Jiang, Liu, Zamir, Toderici, Laptev, Shah, and
  Sukthankar]{jiang2014thumos}
Y.~Jiang, J.~Liu, A.~R. Zamir, G.~Toderici, I.~Laptev, M.~Shah, and
  R.~Sukthankar, ``Thumos challenge: Action recognition with a large number of
  classes,'' 2014.

\bibitem[Weinzaepfel et~al.(2017)Weinzaepfel, Martin, and
  Schmid]{weinzaepfel2017human}
P.~Weinzaepfel, X.~Martin, and C.~Schmid, ``Human action localization with
  sparse spatial supervision,'' \emph{arXiv preprint arXiv:1605.05197}, 2017.

\bibitem[Peng and Schmid(2016)]{peng2016multi}
X.~Peng and C.~Schmid, ``Multi-region two-stream r-cnn for action detection,''
  in \emph{European Conference on Computer Vision}.\hskip 1em plus 0.5em minus
  0.4em\relax Springer, 2016, pp. 744--759.

\bibitem[Kl{\"a}ser(2010)]{klaser2010learning}
A.~Kl{\"a}ser, ``Learning human actions in video,'' Ph.D. dissertation, PhD
  thesis, Universit{\'e} de Grenoble, 2010.

\bibitem[Yu and Joachims(2009)]{yu2009learning}
C.-N.~J. Yu and T.~Joachims, ``Learning structural svms with latent
  variables,'' in \emph{Proceedings of the 26th annual international conference
  on machine learning}.\hskip 1em plus 0.5em minus 0.4em\relax ACM, 2009, pp.
  1169--1176.

\bibitem[Patron-Perez et~al.(2010)Patron-Perez, Marszalek, Zisserman, and
  Reid]{patron2010high}
A.~Patron-Perez, M.~Marszalek, A.~Zisserman, and I.~D. Reid, ``High five:
  Recognising human interactions in tv shows.'' in \emph{BMVC}, vol.~1.\hskip
  1em plus 0.5em minus 0.4em\relax Citeseer, 2010, p.~2.

\bibitem[Leordeanu et~al.(2012)Leordeanu, Sukthankar, and
  Sminchisescu]{leordeanu2012efficient}
M.~Leordeanu, R.~Sukthankar, and C.~Sminchisescu, ``Efficient closed-form
  solution to generalized boundary detection,'' in \emph{European Conference on
  Computer Vision}.\hskip 1em plus 0.5em minus 0.4em\relax Springer, 2012, pp.
  516--529.

\bibitem[Arbelaez et~al.(2009)Arbelaez, Maire, Fowlkes, and
  Malik]{arbelaez2009contours}
P.~Arbelaez, M.~Maire, C.~Fowlkes, and J.~Malik, ``From contours to regions: An
  empirical evaluation,'' 2009.

\bibitem[Tran and Yuan(2011)]{tran2011optimal}
D.~Tran and J.~Yuan, ``Optimal spatio-temporal path discovery for video event
  detection,'' in \emph{Computer Vision and Pattern Recognition (CVPR), 2011
  IEEE Conference on}.\hskip 1em plus 0.5em minus 0.4em\relax IEEE, 2011, pp.
  3321--3328.

\bibitem[Tran and Yuan(2012)]{tran2012max}
------, ``Max-margin structured output regression for spatio-temporal action
  localization,'' in \emph{Advances in neural information processing systems},
  2012, pp. 350--358.

\bibitem[Dalal and Triggs(2005)]{dalal2005histograms}
N.~Dalal and B.~Triggs, ``Histograms of oriented gradients for human
  detection,'' in \emph{Computer Vision and Pattern Recognition, 2005. CVPR
  2005. IEEE Computer Society Conference on}, vol.~1.\hskip 1em plus 0.5em
  minus 0.4em\relax IEEE, 2005, pp. 886--893.

\bibitem[Blank et~al.(2005)Blank, Gorelick, Shechtman, Irani, and
  Basri]{ActionsAsSpaceTimeShapes_iccv05}
M.~Blank, L.~Gorelick, E.~Shechtman, M.~Irani, and R.~Basri, ``Actions as
  space-time shapes,'' in \emph{The Tenth IEEE International Conference on
  Computer Vision (ICCV'05)}, 2005, pp. 1395--1402.

\bibitem[Raptis et~al.(2012)Raptis, Kokkinos, and
  Soatto]{raptis2012discovering}
M.~Raptis, I.~Kokkinos, and S.~Soatto, ``Discovering discriminative action
  parts from mid-level video representations,'' in \emph{Computer Vision and
  Pattern Recognition (CVPR), 2012 IEEE Conference on}.\hskip 1em plus 0.5em
  minus 0.4em\relax IEEE, 2012, pp. 1242--1249.

\bibitem[Wang et~al.(2014)Wang, Qiao, and Tang]{wang2014video}
L.~Wang, Y.~Qiao, and X.~Tang, ``Video action detection with relational
  dynamic-poselets,'' in \emph{European Conference on Computer Vision}.\hskip
  1em plus 0.5em minus 0.4em\relax Springer, 2014, pp. 565--580.

\bibitem[Kipf and Welling(2016)]{kipf2016semi}
T.~N. Kipf and M.~Welling, ``Semi-supervised classification with graph
  convolutional networks,'' \emph{arXiv preprint arXiv:1609.02907}, 2016.

\bibitem[Kay et~al.(2017)Kay, Carreira, Simonyan, Zhang, Hillier,
  Vijayanarasimhan, Viola, Green, Back, Natsev, et~al.]{kay2017kinetics}
W.~Kay, J.~Carreira, K.~Simonyan, B.~Zhang, C.~Hillier, S.~Vijayanarasimhan,
  F.~Viola, T.~Green, T.~Back, P.~Natsev \emph{et~al.}, ``The kinetics human
  action video dataset,'' \emph{arXiv preprint arXiv:1705.06950}, 2017.

\bibitem[Shahroudy et~al.(2016)Shahroudy, Liu, Ng, and Wang]{shahroudy2016ntu}
A.~Shahroudy, J.~Liu, T.-T. Ng, and G.~Wang, ``Ntu rgb+ d: A large scale
  dataset for 3d human activity analysis,'' in \emph{Proceedings of the IEEE
  conference on computer vision and pattern recognition}, 2016, pp. 1010--1019.

\bibitem[Fernando et~al.(2015)Fernando, Gavves, Oramas, Ghodrati, and
  Tuytelaars]{fernando2015modeling}
B.~Fernando, E.~Gavves, J.~M. Oramas, A.~Ghodrati, and T.~Tuytelaars,
  ``Modeling video evolution for action recognition,'' in \emph{Proceedings of
  the IEEE Conference on Computer Vision and Pattern Recognition}, 2015, pp.
  5378--5387.

\bibitem[Liu et~al.(2016)Liu, Shahroudy, Xu, and Wang]{liu2016spatio}
J.~Liu, A.~Shahroudy, D.~Xu, and G.~Wang, ``Spatio-temporal lstm with trust
  gates for 3d human action recognition,'' in \emph{European Conference on
  Computer Vision}.\hskip 1em plus 0.5em minus 0.4em\relax Springer, 2016, pp.
  816--833.

\bibitem[Kim and Reiter(2017)]{kim2017interpretable}
T.~S. Kim and A.~Reiter, ``Interpretable 3d human action analysis with temporal
  convolutional networks,'' in \emph{2017 IEEE Conference on Computer Vision
  and Pattern Recognition Workshops (CVPRW)}.\hskip 1em plus 0.5em minus
  0.4em\relax IEEE, 2017, pp. 1623--1631.

\bibitem[Tran et~al.(2017)Tran, Wang, Torresani, Ray, LeCun, and
  Paluri]{tran2017closer}
D.~Tran, H.~Wang, L.~Torresani, J.~Ray, Y.~LeCun, and M.~Paluri, ``A closer
  look at spatiotemporal convolutions for action recognition,'' \emph{arXiv
  preprint arXiv:1711.11248}, 2017.

\bibitem[Karpathy et~al.(2014)Karpathy, Toderici, Shetty, Leung, Sukthankar,
  and Fei-Fei]{karpathy2014large}
A.~Karpathy, G.~Toderici, S.~Shetty, T.~Leung, R.~Sukthankar, and L.~Fei-Fei,
  ``Large-scale video classification with convolutional neural networks,'' in
  \emph{Proceedings of the IEEE conference on Computer Vision and Pattern
  Recognition}, 2014, pp. 1725--1732.

\bibitem[Kuehne et~al.(2013)Kuehne, Jhuang, Stiefelhagen, and
  Serre]{kuehne2013hmdb51}
H.~Kuehne, H.~Jhuang, R.~Stiefelhagen, and T.~Serre, ``Hmdb51: A large video
  database for human motion recognition,'' in \emph{High Performance Computing
  in Science and Engineering ‘12}.\hskip 1em plus 0.5em minus 0.4em\relax
  Springer, 2013, pp. 571--582.

\bibitem[Caba~Heilbron et~al.(2015)Caba~Heilbron, Escorcia, Ghanem, and
  Carlos~Niebles]{caba2015activitynet}
F.~Caba~Heilbron, V.~Escorcia, B.~Ghanem, and J.~Carlos~Niebles, ``Activitynet:
  A large-scale video benchmark for human activity understanding,'' in
  \emph{Proceedings of the IEEE Conference on Computer Vision and Pattern
  Recognition}, 2015, pp. 961--970.

\end{thebibliography}
}

\end{document}